\documentclass[lettersize,journal]{IEEEtran}
\usepackage{amsmath,amsfonts,bm}
\usepackage{algorithmic}
\usepackage{algorithm}
\usepackage{array}
\usepackage[caption=false]{subfig}
\usepackage{textcomp}
\usepackage{stfloats}
\usepackage{url}
\usepackage{verbatim}
\usepackage{graphicx}
\usepackage{cite}
\usepackage{multirow}
\usepackage{array}
\usepackage{makecell}
\usepackage{float}
\usepackage{stfloats}
\usepackage{color}
\usepackage{tabularray}
\usepackage{booktabs}
\usepackage{bbding}
\usepackage{array}
\usepackage[misc]{ifsym} 
\usepackage{setspace}
\usepackage{fancyhdr}

\begin{document}
\fancyhead[R]{\footnotesize \thepage}
\markboth{\footnotesize \textsc{IEEE Transactions on Image Processing}}{\footnotesize \textsc{IEEE Transactions on Image Processing}}
\twocolumn
\title{Unsupervised Modality-Transferable Video Highlight Detection with Representation Activation Sequence Learning}
% \markboth{\footnotesize \textsc{IEEE Transactions on Image Processing}}{\footnotesize \textsc{IEEE Transactions on Image Processing}}

\author{Tingtian Li\textsuperscript{1*}, Zixun Sun\textsuperscript{1}, and Xinyu Xiao\textsuperscript{1}

% <-this % stops a space

\thanks{\textsuperscript{1}Tencent, China.}
\thanks{\textsuperscript{*}Email: tingtian.li@outlook.com.}
}

\maketitle

\begin{abstract}
Identifying highlight moments of raw video materials is crucial for improving the efficiency of editing videos that are pervasive on internet platforms. However, the extensive work of manually labeling footage has created obstacles to applying supervised methods to videos of unseen categories. The absence of an audio modality that contains valuable cues for highlight detection in many videos also makes it difficult to use multimodal strategies. In this paper, we propose a novel model with cross-modal perception for unsupervised highlight detection. The proposed model learns representations with visual-audio level semantics from image-audio pair data via a self-reconstruction task. To achieve unsupervised highlight detection, we investigate the latent representations of the network and propose the representation activation sequence learning (RASL) module with {\itshape k}-point contrastive learning to learn significant representation activations. To connect the visual modality with the audio modality, we use the symmetric contrastive learning (SCL) module to learn the paired visual and audio representations. Furthermore, an auxiliary task of masked feature vector sequence (FVS) reconstruction is simultaneously conducted during pretraining for representation enhancement. During inference, the cross-modal pretrained model can generate representations with paired visual-audio semantics given only the visual modality. The RASL module is used to output the highlight scores. The experimental results show that the proposed framework achieves superior performance compared to other state-of-the-art approaches.
\end{abstract}

\begin{IEEEkeywords}
unsupervised, video highlight detection, multimodal, representation activation sequence.
\end{IEEEkeywords}

\section{Introduction}
\IEEEPARstart{W}{ith} the prevalence of video media on internet platforms, there is a growing demand for automatically extracting highlight segments from sheer volumes of footage for quick browsing. Especially in the era of flourishing short videos, fast production means both timely and reducing labor costs. Thus, as a technique that automatically locates attractive segments, highlight detection has attracted increasing attention from researchers in recent years \cite{gygli2016video2gif,yang2015unsupervised,sun2014ranking,xiong2019less,yao2016highlight,liu2015multi,jiao2017video,rochan2020adaptive,ye2021temporal,hong2020mini,sharghi2016query,zeng2019graph,gan2015devnet}.

\begin{figure}[h]
  \centering
  \includegraphics[width=7cm]{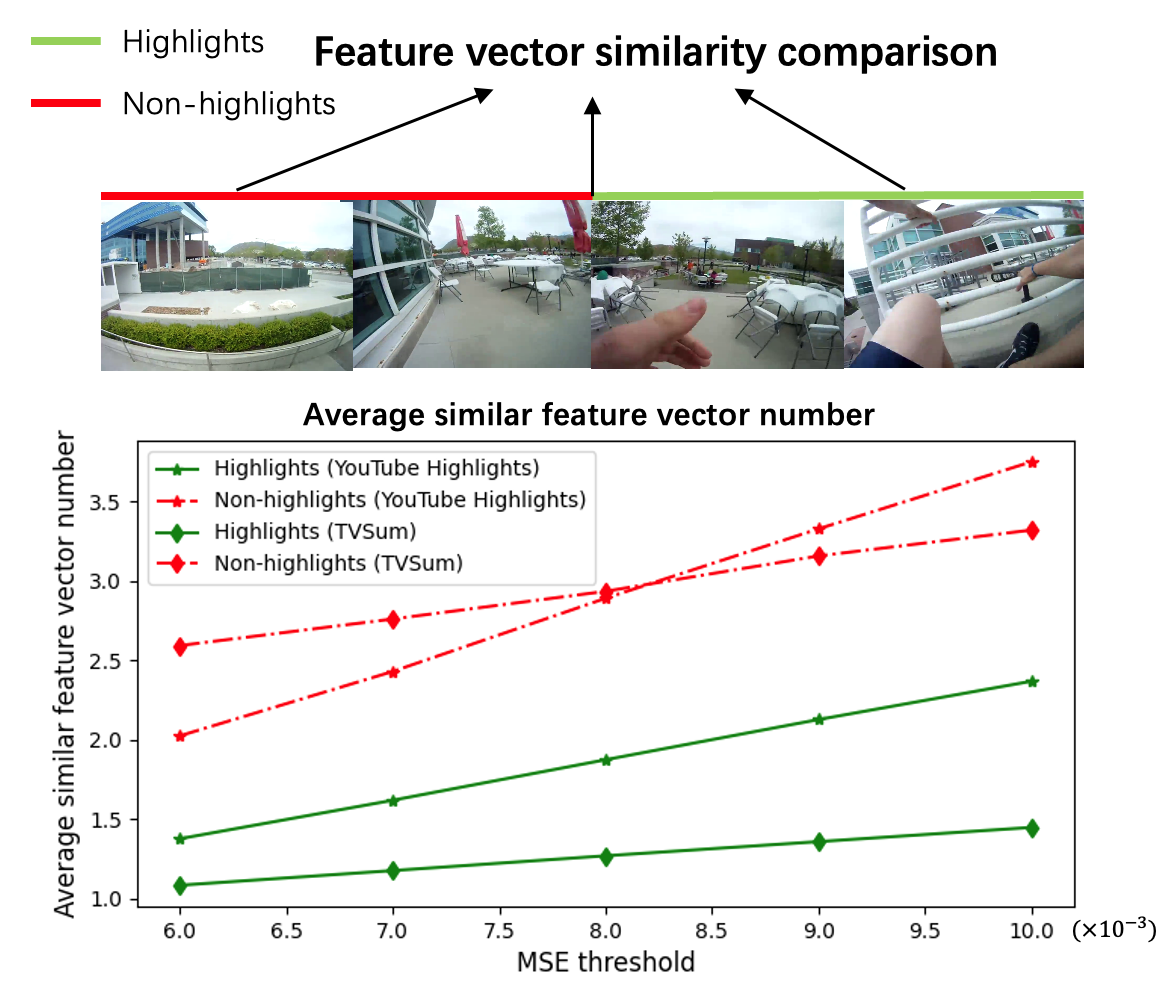}
  \caption{We evaluate the similarity of feature vectors extracted by the feature extractor \cite{liu2021swin} from videos in the YouTube Highlights and TVSum datasets and show the number of feature vectors that exhibit similarity to others within the videos. We calculate the mean squared error (MSE) to assess the similarity between the target feature vector and others within a window (30 seconds). The sampling rate is set to 5 frames/second. If the MSE is below the threshold, we consider the two vectors to be similar. The results show that highlights have fewer similar feature vectors.
 }
\end{figure}

The majority of highlight detection approaches are supervised \cite{gygli2016video2gif,sun2014ranking,jiao2017video,rochan2020adaptive}. Given videos with frame-level labels, supervised methods learn from frame features and temporally localize highlight segments. However, as frame-level labeling is labor intensive, datasets with frame-level labels are unable to include vast video categories. Supervised approaches thus become domain specific and show a weak ability to adapt to wild videos of unseen categories. Another branch of methods corresponds to weakly supervised methods \cite{yang2015unsupervised,xiong2019less,ye2021temporal,hong2020mini}. Weakly supervised methods do not rely on expensive frame-level labels. Usually, they are heuristic and learn from specific metadata, e.g., video duration \cite{xiong2019less} and video categories \cite{yang2015unsupervised,ye2021temporal,hong2020mini}. However, metadata still require special preparation and result in weak generalization.
The association of visual and audio modalities has been employed in different applications  \cite{sanguineti2021audio
,akbari2021vatt,li2021deep,afouras2018deep}. The methods \cite{ye2021temporal,hong2020mini} also aware that the audio modality is useful for highlight detection. However, the audio modality is unavailable in some situations. For example, a camera may be far from the sound source of an attractive scene, or the original sound may be obscured by irrelevant environmental noise or background music.

\begin{figure}[t]
  \flushleft 
  \subfloat[The training and inference flowchart of the weakly supervised visual-audio-based methods \cite{ye2021temporal,hong2020mini}]{\includegraphics[width=3.3in]{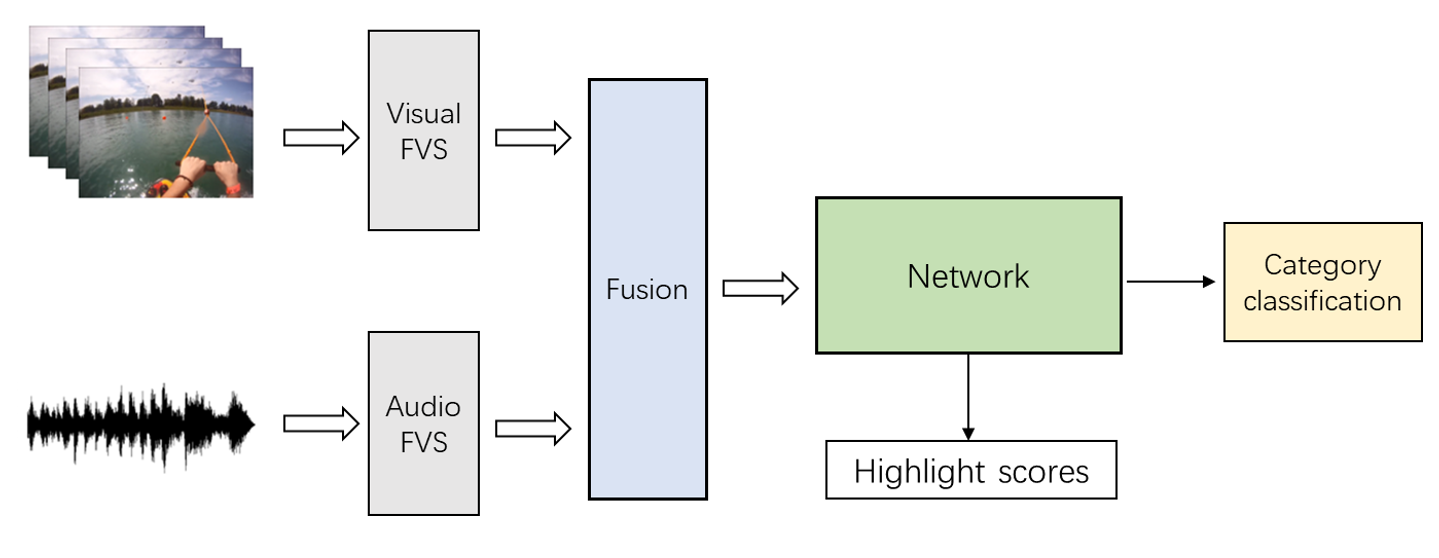}} \\
  \subfloat[The pretraining flowchart of the proposed unsupervised method]{\includegraphics[width=3.5in]{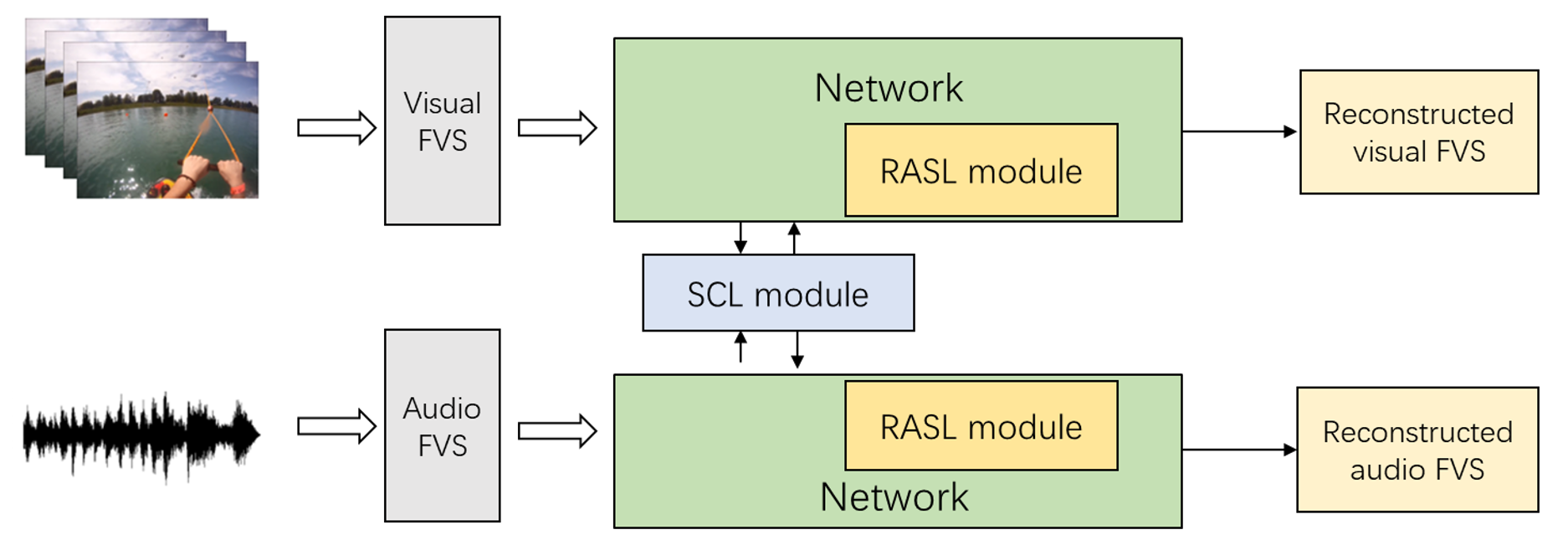}}\\
  \subfloat[The inference flowchart of the proposed unsupervised method]{\includegraphics[width=3.4in]{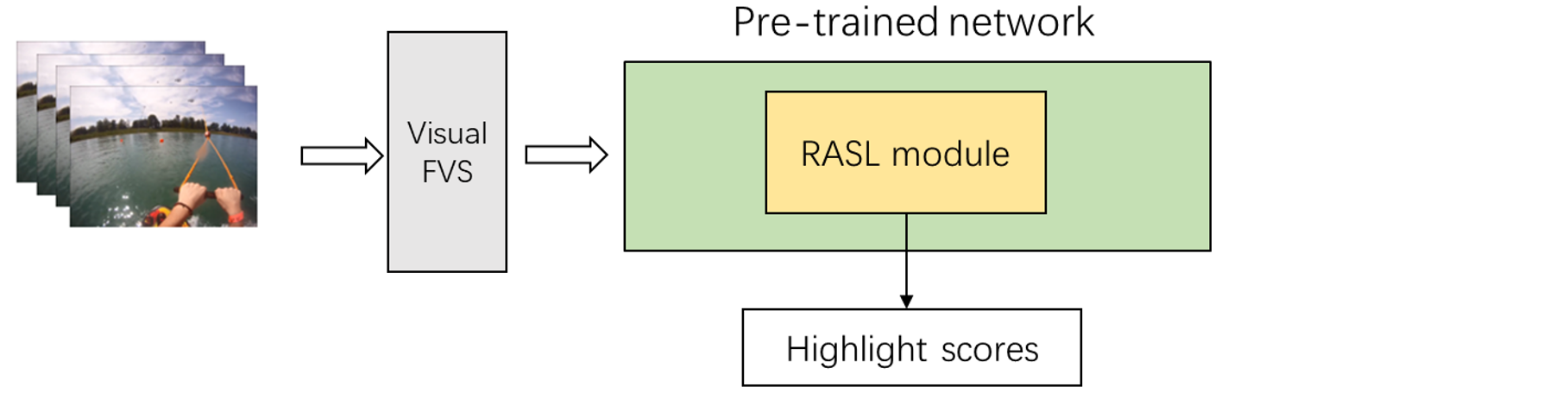}}
  \caption{Flowcharts of the weakly supervised visual-audio-based methods \cite{ye2021temporal,hong2020mini} and the proposed unsupervised method. \cite{ye2021temporal,hong2020mini} need both the visual and the audio modalities as the input in the training and inference processes. The proposed method requires the visual and audio modalities as the input during pretraining. During inference, only the visual modality is needed to obtain representations with visual-audio-level semantics from the pretrained model, and the RASL module is used to output the highlight scores.}
\end{figure}
\IEEEpubidadjcol

In this paper, we propose a novel unsupervised cross-modal highlight detection framework. Motivated by the saliency detection task, where salient regions that attract our attention in an image are distinct from its overall appearance\cite{cheng2014global, wang2016background}, we consider highlight frames, which attract our attention, to be highly likely to exhibit semantic distinctiveness in the footage compared to ordinary frames. To verify this, we analyze the similarity of feature vectors to others within the videos in the YouTube Highlights \cite{sun2014ranking} and TVSum \cite{song2015tvsum} datasets. We present the average numbers of similar feature vectors to others in Fig. 1. The results show that highlights have fewer similar feature vectors, indicating that they are more distinct than non-highlights are. From the view of information theory, we argue that highlights contain more meaningful information than ordinary snippets because they are more distinct. Inspired by this, we construct a network with temporal activations and propose the representation activation sequence learning (RASL) module to learn the significant representation activations through self-reconstruction during pretraining. The original purpose of training deep neural networks for signal reconstruction was to solve inverse problems \cite{jin2017deep, zhang2017beyond, li2020improved}. Recently, self-reconstruction has become frequently used in self-supervised model pretraining for improving the performance of downstream tasks, such as image and video classification, object detection, and semantic segmentation \cite{dosovitskiy2020image,he2022masked,baevski2020wav2vec, zhuang2020unsupervised,chen2021pre}. However, the focus of these works is primarily on obtaining effective representations from intermediate layers of pretrained networks for downstream tasks, and the use of representation activations of self-reconstruction networks for direct application is often overlooked. In this study, we detect highlight moments using the representation activations of the pretrained network in a self-supervised manner. The proposed RASL module promotes larger values in the representation activations for distinct highlight snippets for better reconstruction, as they are more challenging to reconstruct than redundant ordinary frames are. This enables the model to infer highlight moments from the representation activation sequence. Specifically, the RASL module aggregates the top-{\itshape k} representation activations with the highest responses and guides them to be more distinguishable. As the activations selected by the fixed hyperparameter {\itshape k} probably incorporate outliers, we propose {\itshape k}-point contrastive learning to suppress the outliers.
The framework contains two branches of autoencoders operating on the visual and audio modalities. We use the symmetric contrastive learning (SCL) module to establish the connection between the paired visual and the audio representations via modal contrastive learning \cite{radford2021learning}. During inference, the visual branch can generate representations with paired visual-audio semantics and output highlight scores via the RASL module without the need for the audio modality.
Figure 2 shows the flowcharts of the weakly supervised visual-audio methods \cite{ye2021temporal,hong2020mini} and the proposed unsupervised method. \cite{ye2021temporal,hong2020mini} require both visual frames and audio waves as inputs during training and inference. In contrast, the proposed method requires both modalities only during pretraining. During inference, given only the visual frames, the pretrained cross-modal network and the RASL module are used to output the highlight scores. To our knowledge, the proposed method is the first visual-audio-based method that does not require any audio as the inference input. Note that weakly supervised multimodal approaches \cite{ye2021temporal,hong2020mini} also need video category labels, whereas the pretraining of the proposed methods is unsupervised without the need for any labels or metadata.
We also utilize multitask learning (MTL) to improve the framework performance. MTL is an approach for improving the performance of a network by using the domain information contained in another simultaneously trained task. Here, we adopt masked feature vector sequence (FVS) reconstruction, which has been shown to improve representation performance in self-supervised learning studies \cite{he2022masked,baevski2020wav2vec,devlin2018bert}, as an auxiliary task.

To summarize, the main contributions of our proposed method are as follows:

1) We propose a novel unsupervised cross-modal highlight detection framework. We use a self-reconstruction task to pretrain the model and build the visual-audio connection. Given only visual frames, the pretrained model can generate representations with cross-modal semantics according to the modal connection. It can directly infer highlights of wild videos without further training.

2) We propose the RASL module with {\itshape k}-point contrastive learning to guide the activations of significant representations to be more distinguishable and suppress the incorporated activation outliers without the need for frame-level annotated labels. During inference, the RASL module outputs the highlight scores from the learned representations.

3) We use the SCL module to build the modal connection with cross-modal contrastive learning. The SCL module enables the visual branch to generate representations with visual-audio-level semantics from learning to pair representations.

4) We build an MTL framework and add an auxiliary task of masked FVS reconstruction. The auxiliary task enhances the representations and improves the highlight detection performance.

\begin{figure*}[t]
  \centering
  \includegraphics[width=15.5cm]{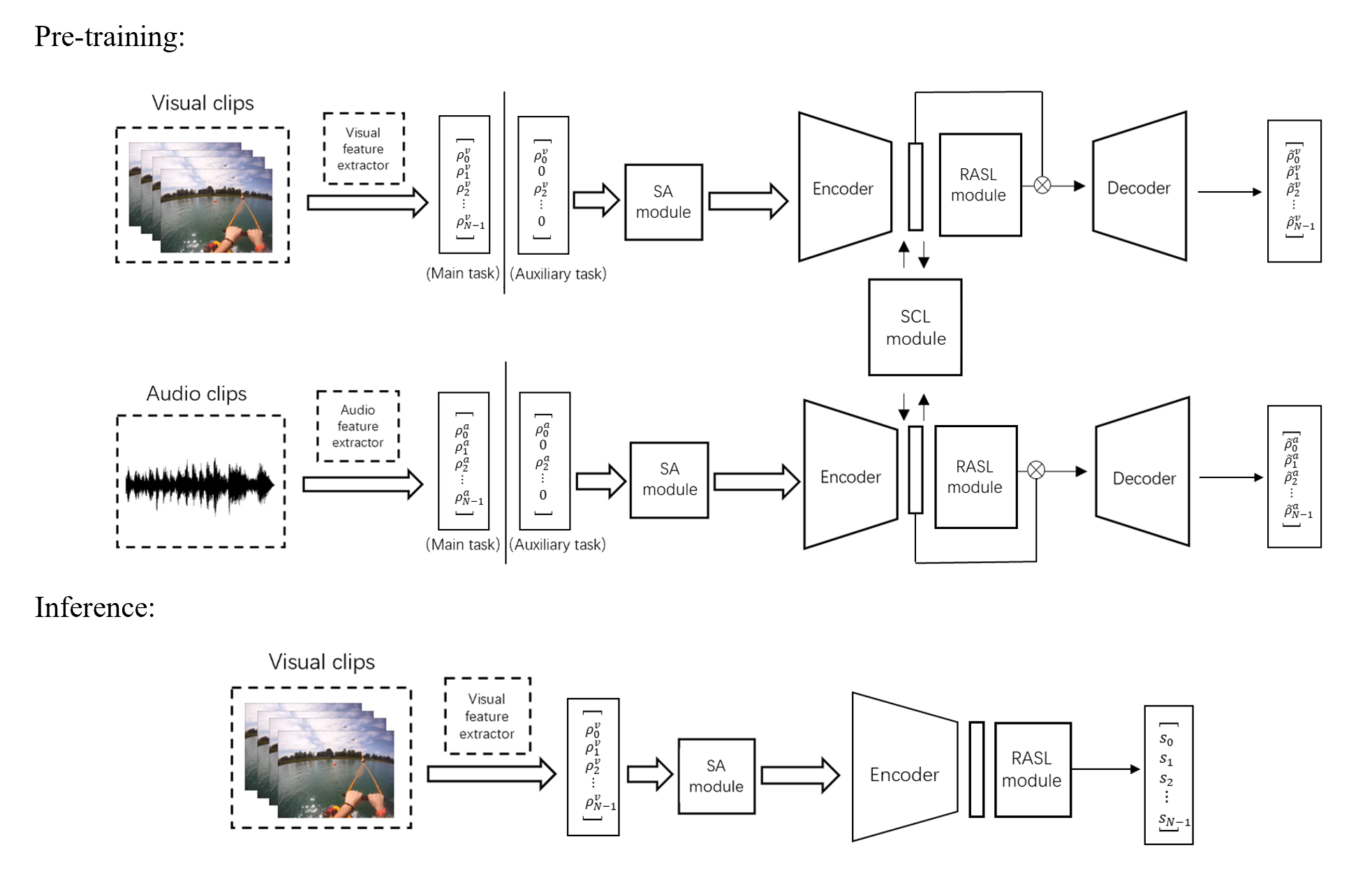}
  \caption{The framework of the proposed highlight detection method. During pretraining, we input the visual and audio clips into the two branches and extract their feature vectors to compose the visual and audio FVSs. Then, we enhance the FVSs of the two modalities via the SA modules. After that, the self-attended FVSs are fed to the autoencoders for self-reconstruction. The significant activations are learned from the RASL module. The paired visual-audio representations are learned through the SCL module. The auxiliary task of masked FVS reconstruction is conducted to improve the performance of the main highlight detection task. During inference, we use only the cross-modal pretrained visual branch and the RASL module to output highlight scores.}
\end{figure*}

\section{Related Work}
\subsection{Video Highlight Detection}
% \noindent {\bf Video Highlight Detection:}
Early highlight detection works focused on solving the sporting video clipping problem \cite{tang2011detecting,wang2004sports,xiong2005highlights}. Some later works localized attractive segments for social media production \cite{sun2014ranking,yao2016highlight}. \cite{sun2014ranking,yao2016highlight,jiao2017video,yu2018deep,wang2020learning,gygli2016video2gif} formulated attractive segment localization as a segment ranking problem. They trained ranking networks using contrastive learning and assigned higher scores to highlight segments. These methods are supervised and require the availability of frame-level annotated labels. However, manually labeling highlight moments from footage is labor intensive. As a result, the number of annotated videos is limited, and supervised methods easily fall into domain-specific problems. To avoid the limited frame-level annotation problem, weakly supervised highlight detection approaches have been studied \cite{yang2015unsupervised,xiong2019less,ye2021temporal,hong2020mini}. Without frame-level annotation, weakly supervised approaches detect highlight segments with the assistance of metadata. \cite{yang2015unsupervised} distinguished highlight frames from FVS reconstruction errors of specific video categories. \cite{xiong2019less} demonstrated that videos of shorter duration have higher probabilities of being highlight clips. This fact implicitly supervises the network to prefer segments from shorter videos. \cite{ye2021temporal,hong2020mini} reported that the audio modality can assist in detecting highlight snippets. \cite{hong2020mini} proposed to adopt a ranking loss and uses multiple submodules to fuse the modalities. \cite{ye2021temporal} utilized a hierarchical temporal encoder and a multimodal tensor fusion mechanism to fuse modalities. However, existing visual-audio-based methods \cite{ye2021temporal,hong2020mini} require visual frames and audio waves for both training and inference. They cannot handle situations where the original sound is lost or strongly disturbed. Moreover, although \cite{yang2015unsupervised,xiong2019less,ye2021temporal,hong2020mini} avoid frame-level annotation, they still need to prepare the dataset with metadata in specific ways, e.g., video topics \cite{yang2015unsupervised,ye2021temporal,hong2020mini} and annotations of video durations \cite{xiong2019less}. In contrast, our proposed method learns from the unsupervised self-reconstruction task without requiring any metadata. During inference, the proposed method requires only visual frames but can generate representations with learned visual-audio semantics to infer the highlight segments of videos in the wild.

\subsection{Video Summarization}

Video summarization is a task similar to video highlight detection. The goal is to integrate the important frames and generate a compact summary of a given video \cite{mahasseni2017unsupervised,kanafani2021unsupervised,yuan2019cycle,chu2015video,cai2018weakly,elhamifar2012see,kim2014joint,potapov2014category,gong2014diverse,zhang2016video,panda2017weakly,panda2017collaborative}. \cite{potapov2014category} proposed to learn the summaries from category-specific videos. \cite{zhang2016video} detected important frames using a probabilistic model for diverse sequential subset selection. \cite{gong2014diverse} utilized long short-term memory (LSTM) \cite{graves2012long} to model the variable-range dependencies for video summarization. \cite{chu2015video} located the segments that cooccur most frequently across collected videos using a keyword. \cite{cai2018weakly} learned semantic matching between the generated summaries and web videos. Several other methods \cite{mahasseni2017unsupervised,kanafani2021unsupervised,yuan2019cycle} utilized the generative adversarial network (GAN) \cite{radford2016unsupervised,arjovsky2017wasserstein,li2019image} to regularize the summarizer by validating the consistency between the estimated summaries and the video features.

\subsection{Multitask Learning and Masked Signal Reconstruction}
The MTL paradigm aims to leverage useful knowledge contained in multiple related tasks that are trained simultaneously to improve the performance and data efficacy of all the tasks \cite{baxter1997bayesian,duong2015low,yang2016trace}. MTL is similar to transfer learning \cite{weiss2016survey} in that it utilizes the information in pretrained tasks to improve the performance of downstream tasks. However, MTL trains multiple tasks simultaneously, leveraging their shared knowledge to improve model generalizability and performance. In this work, we adopt the MTL strategy for performance improvement.

Masked signal reconstruction, as a popular task of self-supervised learning, has been applied to different modal signals, e.g., vision \cite{dosovitskiy2020image,he2022masked}, language \cite{devlin2018bert}, and audio \cite{baevski2020wav2vec,liu2020mockingjay}. This approach predicts the original signals from these intentional distortions and shares the learned knowledge with other tasks. Specifically, \cite{he2022masked} demonstrated that autoencoders with downsampling and upsampling abilities are naturally suitable for self-supervised reconstruction of mask signals. Considering that the proposed framework also contains an autoencoder structure, we select masked FVS reconstruction as our auxiliary task.

\begin{figure*}[t]
  \centering
  \includegraphics[width=14.8cm]{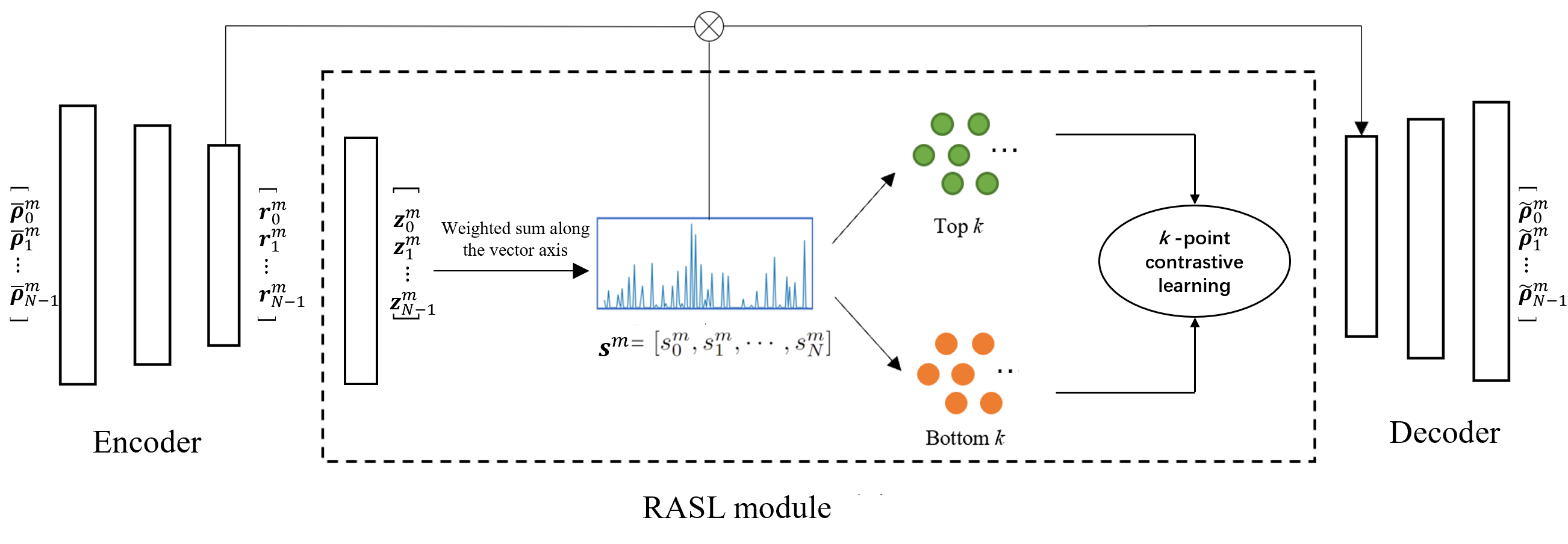}
  \caption{The illustration of the proposed RASL module, which is shown in the dotted box. The representation vectors $\bm{{r}}^{m}$ output from the encoder are sent to the module, which learns $\bm{{s}}^{m}$ via {\itshape k}-point contrastive learning. Then, the product of $\bm{{s}}^{m}$ and $\bm{{z}}^{m}$ is fed to the decoder for FVS reconstruction.
 }
\end{figure*}

\section{Method}
Fig. 3 shows the framework of our proposed method. During pretraining, we extract the FVSs of the visual and audio modalities and feed them into the autoencoders with the RASL modules for self-reconstruction. The representations with visual-audio semantics are learned by the SCL module. The auxiliary task of masked FVS reconstruction is used to improve the highlight detection performance. During inference, we use the cross-modal pretrained visual branch and the RASL module to output the highlight scores. In this section, we will illustrate the details of the proposed method as follows.

\subsection{Representations of The Visual and The Audio Modalities}
Our method learns from visual and audio representations to detect video highlights during pretraining. We first transform the raw visual and audio signals into FVSs, which are also the targets of self-reconstruction. Here, we denote the visual and the audio modalities as $v$ and $a$, respectively. For the visual modality $v$, we first extract the frames of the target video at a fixed interval, resulting in $N$ frames. We map the extracted {\itshape i}-th frame to the feature vector $ \bm{\rho}_{i}^{v} \in{R^{1\times{d_{v}}}} $ by the feature extractor \cite{liu2021swin}, where $d_{v}$ is the visual feature vector dimension. We denote the visual FVS as $ \bm{\rho}^{v}= [\bm{\rho}^{v}_{0},\bm{\rho}^{v}_{1},\cdots,\bm{\rho}^{v}_{N-1}] $. For the audio modality $a$, we extract the feature vectors using the feature extractor \cite{liu2020mockingjay}. Since the audio signal temporal density is much greater than the visual signal temporal density, we further downsample the audio FVS via average pooling for network parameter reduction. We denote the obtained audio FVS as $ \bm{\rho}^{a} = [\bm{\rho}^{a}_{0},\bm{\rho}^{a}_{1},\cdots,\bm{\rho}^{a}_{N-1}] $, where $\bm{\rho}^{a}_{i} \in{R^{1\times{d_{a}}}} $ indicates the $i$-th audio feature vector with dimension $d_{a}$.

\subsection{Intramodal Self-Attention}
Highlighting moments within a video are not independent. They depend on the video structure of the contents and are semantically related. Here, we use the self-attention (SA) module \cite{vaswani2017attention} to model this relationship in our representation generation. The SA module learns temporal weights according to the feature vectors at other temporal locations. It effectively models the temporal dependency of the feature vectors. Specifically, given an FVS $ \bm{\rho}^{m}= [\bm\rho_{0}^{m},\bm\rho_{1}^{m},\cdots,\bm\rho_{N-1}^{m}]$, we project $\bm{\rho}^{m}$ to the value sequence $ V^m$, the query sequence $ Q^m $ and the key sequence $ Q^m $, where $\bm{\rho}_{i}^{m} \in{R^{1\times{d_{\rho}^{m}}}} $ is the $i$th feature vector of the modality $m$ with dimensions $d_{\rho}^{m}$ and $m\in{\left\{v, a\right\}}$. We model the dependency by the weight $ V^m $, which is calculated by the multiplication of $ Q^m $ and $ K^m $. The calculation of the SA mechanism is shown as follows:
\begin{equation}
\begin{aligned}
\label{eq1}
O_{\bm{\rho}^{m}}&= softmax(\frac{Q^{m}(K^m)^{T}}{\sqrt{d_{\rho}^{m}}})V^m\\
&=softmax(\frac{\bm{\rho}^{m}W^{Q,m}{(\bm{\rho}^{m}W^{K,m})}^{T}}{\sqrt{d_{\rho}^{m}}})\bm{\rho}^{m}W^{V,m},
\end{aligned}
\end{equation}

\begin{equation}
\label{eq2}
\bar{\bm{\rho}}^{m} = F^m(\bm{\rho}^{m}) = \bm{\rho}^{m} + O_{\bm{\rho}^{m}}W^{O,m},
\end{equation}

\noindent
where $O_{\rho^m}$ is the self-attended FVS and $F^m$ represents the SA process for modality $m$. $W^{Q,m} \in R^{d_{\rho}^{m}\times{d_{\rho}^{m}}}$, $W^{K,m} \in R^{d_{\rho}^{m}\times{d_{\rho}^{m}}}$, $W^{V,m} \in R^{d_{\rho}^{m}\times{d_{\rho}^{m}}}$ and $W^{O,m} \in R^{d_{\rho}^{m}\times{d_{\rho}^{m}}}$ are four learnable projection matrices. $\sqrt{d_{\rho}^{m}}$ is a scaling factor. The temporal dependency of the FVS is modeled by the second part of Eq. (2). We combined $O_{\rho^m}$ with the original FVS $\bm{\rho}^m$ to obtain the enhanced FVS $ \bar{\bm{\rho}}^{m}= [\bar{\bm\rho}_{0}^{m},\bar{\bm\rho}_{1}^{m},\cdots,\bar{\bm\rho}_{N-1}^{m}]$.

\subsection{Representation Activation Selection Learning}
Fig. 1 shows that highlight snippets are often more distinct from overall videos while ordinary frames are more semantically similar and redundant. We argue that highlights contain more meaningful information, as in information theory, the information content of an event is measured by $-log\left(P\left(A\right)\right)$, where the information content increases as the probability $P\left(A\right)$ of an event $A$ decreases. Compared to the redundant ordinary FVS, the highlight FVS with distinct information is more difficult to reconstruct. Inspired by this, we build a network with temporal representation activations for highlight detection. During pretraining, the network tends to learn high activations on distinct representations to minimize the reconstruction error. To guide the activations corresponding to highlight moments to be more distinguishable, we propose the RASL module, which learns larger activations on the significant representation vectors. Thus, we can recognize the highlight moments via the values of representation activations during inference. The structure of the RASL module is shown in Fig. 4. The module learns the representation activation sequence $ \bm{{s}}^{m} = [s^{m}_{0},s^{m}_{1},\cdots,s^{m}_{N-1}] $ of modality $m$ from the representation vector sequence $\bm{r}^{m} $. The representation vector sequence is obtained by
\begin{equation}
\begin{aligned}
\label{eq3}
\bm{r}^{m} = [\bm{r}^{m}_{0},\bm{r}^{m}_{1},\cdots,\bm{r}^{m}_{N-1}] = E^m(\bar{\bm{\rho}}^{m}) = E(F^m({\bm{\rho}}^{m})),
\end{aligned}
\end{equation}

\noindent 
where $E^{m}$ is the encoder of the autoencoder of modality $m$. The {\itshape i}-th activation for modality $m$ is calculated as follows:

\begin{equation}
\begin{aligned}
\label{eq4}
{s}^{m}_{i} &= \sum_{c} w_c^m \bm{z}^{m}_{i}\left[c\right]\\
&= \sum_{c}  w_c^m \psi^m \left( \bm{r}^{m}_{i}\right)\left[c\right] ,
\end{aligned}
\end{equation}

\noindent
where $\psi^m$ represents a convolutional layer. $\bm{z}^{m}_{i} = \psi^m \left( \bm{r}^{m}_{i}\right)$ is the representation vector processed by the convolutional layer $\psi^m$. The representation activation ${s}^{m}_{i}$ is obtained by the weighted sum of $\bm{z}^{m}_{i}$ along the vector axis. $c$ is the index of the vector axis. $w_c^m$ is the result of the weighted sum process. We implement the weighted sum process using a bias-free fully connected layer, where $w_c^m$ serves as a weight of this layer. The products of the sequences $ \bm{{s}}^{m}$ and $\bm{r}^{m}$ are then sent to the decoder for FVS reconstruction. We reconstruct the FVS of modality $m$ by minimizing the reconstruction loss
\begin{equation}
\begin{aligned}
\label{eq5}
L_{e}^m &=  \left | \ \bm{\rho}^{m}-   \tilde{\bm{\rho}}^{m} \right | \\
& = \left | \ \bm{\rho}^{m}- D^{m}( \bm{{s}}^{m}  \cdot \bm{r}^{m} ) \right |\\
& = \left | \ \bm{\rho}^{m}- D^{m}\left([s^{m}_{0} \cdot \bm{r}^{m}_{0},s^{m}_{1} \cdot \bm{r}^{m}_{1},\cdots,s^{m}_{N-1} \cdot \bm{r}^{m}_{N-1}] \right) \right |,
\end{aligned}
\end{equation}

\noindent 
where $D^{m}$ is the decoder of modality $m$. $\tilde{\bm{\rho}}^{m} $ is the reconstructed FVS of modality $m$. In Eq. (5), the representation activation ${s}^{m}_{i}$ serves as an attention weight for $\bm{r}^{m}_{i}$. Therefore, $\bm{s}^{m}$ can be used to denote the highlight temporal locations and is regarded as the highlight score sequence in inference. We also guide $ \bm{{s}}^{m}$ to be more distinguishable at the highlight moments. We use the top-{\itshape k} pooling to aggregate the {\itshape k} activations with the highest probabilities and enlarge their values. The sorted top-{\itshape k} activation index is obtained as

\begin{equation}
\begin{aligned}
\label{eq6}
\Phi^k = \{i \in &\{0, \dots,  k-1  \} \big|l_i \in \Omega^k \text{ and }\\
&{s}^{m}_{l_i} \leq {s}^{m}_{l_{i+1}}\text{  }for\text{  }i<k-1 \}, \\
\text{s.t. } \Omega^{k} &= \mathop{arg max}\limits_{\Gamma\subseteq{\left\{1,\dots,N\right\}}, \left|\Gamma\right|=k} \sum_{i\in\Gamma}{s}^{m}_{i} .
\end{aligned}
\end{equation}

\begin{figure}[t]
  \centering
  \includegraphics[width=8.5cm]{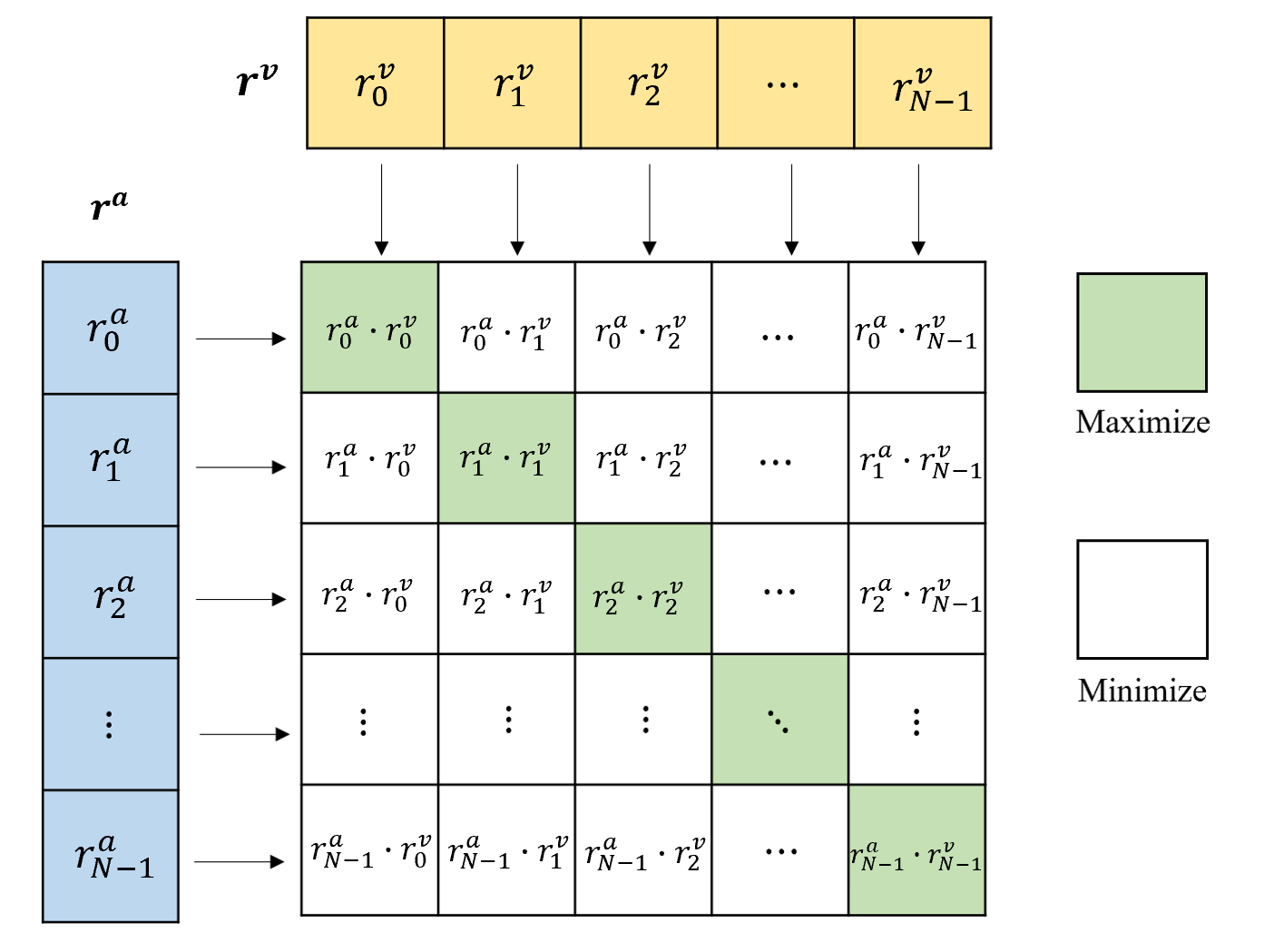}
  \caption{The demonstration of the SCL module. The module maximizes and minimizes the multiplied values of the paired and the unpaired representation vectors, respectively.}
\end{figure}

\noindent 
However, the hyperparameter {\itshape k} is fixed, and we cannot guarantee that all the selected {\itshape k} activations correspond to highlight moments. To solve this noisy selection problem, we propose {\itshape k}-point contrastive learning to suppress the outliers of the selected activations. This mechanism assigns attention weights to the top activations according to the similarity of the top and the bottom processed representation vectors. The sorted bottom-{\itshape k} index set can be obtained as

\begin{equation}
\begin{aligned}
\label{eq7}
\Pi^k = \{i \in &\{0, \dots,  k-1  \} \big|l_i \in \Psi^k \text{ and }\\
&{s}^{m}_{l_i} \leq {s}^{m}_{l_{i+1}} \text{ }for\text{  }i<k-1 \}, \\
\text{s.t. } \Psi^{k} &=\mathop{arg min}\limits_{\Gamma\subseteq{\left\{1,\dots,N\right\}}, \left|\Gamma\right|=k} \sum_{j\in\Gamma}{s}^{m}_{j} .
\end{aligned}
\end{equation}

\noindent 
The objective function of the {\itshape k}-point contrastive learning can be expressed as follows:

\begin{equation}
\begin{aligned}
\label{eq8}
L_{r}^m &= -log{\frac{1}{1+e^{-\eta}}}, \\
\eta  = \frac{1}{k} \sum_{i=0}^{k-1} \hat{s}^{m}_{i} \cdot [ 1-&sim(\sigma(\hat{W_i}\hat{\bm{z}}^{m}_{i}),  \sigma(\check{W_i}\check{\bm{z}}^{m}_{i}) ) ]  ,
\end{aligned}
\end{equation}

\noindent 
where $\hat{s}^{m}_{i} \in \left\{ s^{m}_{i} | i \in \Phi^k \right\}$, $\hat{\bm{z}}^{m}_{i} \in \left\{  \bm{z}^{m}_{i} | i \in \Phi^k \right\}$, $\check{\bm{z}}^{m}_{i} \in \left\{  \bm{z}^{m}_{i} | i \in \Pi^k \right\}$, and $sim$ represent the cosine similarity functions. $\hat{W_i}$ and $\check{W_i}$ are two projection matrices for $\hat{\bm{z}}^{m}_{i}$ and $\check{\bm{z}}^{m}_{i}$, respectively. $\sigma$ denotes the layer normalization function. Through Eq. (8), the top activations $\hat{s}^{m}_{i}$ are enlarged. If $\hat{\bm{z}}^{m}_{i}$ is similar to $\check{\bm{z}}^{m}_{i}$ in the projected space, $\hat{s}^{m}_{i}$ will be assigned a low weight. Thus, the impact of noisy activations is suppressed. When $L_{r}$ is minimized, the distance between $\hat{\bm{z}}^{m}_{i}$ and $\check{\bm{z}}^{m}_{i}$ increases. Through {\itshape k}-point contrastive learning, the top-{\itshape k} activation points for the highlight moments become more distinguishable, while the incorporated noisy outliers are suppressed.

\subsection{Symmetric Contrastive Learning Module}
The visual and audio modalities are highly correlated in highlight detection. For example, when a basketball player has a brilliant goal on the court, there are cheers from the crowd. To build the connection between the two modalities, we use the SCL module, as shown in Fig. 5, to learn the paired representations $\bm{r}^{v}$ and $\bm{r}^{a}$. $\bm{r}^{v}$ and $\bm{r}^{a}$ are the representation vector sequences generated from the encoders of the visual and the audio modalities, respectively. The learned representations with visual-audio-level semantics will continue to be processed separately by the two branch networks. The SCL module is inspired by \cite{radford2021learning}, which uses symmetric contrastive learning to predict the labels of image and text pairs for unsupervised learning. We utilize the SCL module to enable representations to have paired cross-modal semantics. This is achieved by minimizing the term

\begin{equation}
\begin{aligned}
\label{eq9}
L_{s}  = -  \Gamma \left(\sum_{i}{\log{\left((\bm{r}^{a}_{i})^{T}\bm{r}^{v}_{i}\right)}} - \sum_{i}{\sum_{j\not=i}\log{\left( (\bm{r}_{i}^{a})^{T}\bm{r}^{v}_{j}\right)  }}\right),
\end{aligned}
\end{equation}

\noindent 
where $\Gamma $ is a learnable temperature parameter. It serves as a learnable weight that is independent of the network structure and optimized during pretraining. It controls the strength of symmetric contrastive learning, which constructs the connection between the visual and audio modalities. In Eq. (9), we use multiplication to calculate the similarity values of two modal representation vectors. The paired visual and audio representation vectors are associated by maximizing their similarity values, while the similarity values of unpaired representations are suppressed by minimization. Through the SCL module, the representation vectors of the audio and the visual modalities interact. These vectors are then processed separately through their respective branches without interlacing their structures. During inference, a single modality can be input to its corresponding branch using cross-modal pretrained knowledge to generate the highlight score sequence. Here, we adopt the cross-modal pretrained visual branch and use its RASL module to generate the highlight score sequence $ \bm{s}^v$ as our final highlight score sequence $ \bm{s} = [s_{0},s_{1},\cdots,s_{N}]$.

\subsection{The Auxiliary Task of Masked FVS Reconstruction}
As our RASL module highly relies on the representations, improving the efficiency of the representations is essential for unsupervised highlight detection. Unsupervised masked signal reconstruction has shown its ability to improve the latent representation performance for downstream applications \cite{dosovitskiy2020image,he2022masked, vaswani2017attention}. Specifically, \cite{he2022masked} demonstrated that the autoencoder is suitable for the masked signal reconstruction task. Inspired by this, we incorporate an auxiliary masked FVS reconstruction task within the framework. This task leverages the existing autoencoder structure and does not introduce additional complexity to the network. By reconstructing the masked FVS, the model learns to make predictions based on existing video information and understands the video semantics, enabling it to learn meaningful representations and effectively handle scenarios where certain video information is missing. This makes the model more robust and adaptable to real-world scenarios. The training of the auxiliary task relies only on the following reconstruction loss:

\begin{equation}
\begin{aligned}
\label{eq10}
L_{au}^{m} &= \left | \ \bm{\rho}^{m}- D^{m}( \bm{s}^{m} \cdot E^m(F^m({\bm{M} \cdot \bm{\rho}}^{m}))) \right |\\
&= \left | \ \bm{\rho}^{m}- D^{m}( \bm{s}^{m} \cdot E^m(F^m({\dot{\bm{\rho}}}^{m}))) \right |,
\end{aligned}
\end{equation}

\noindent
where $ \dot{\bm{\rho}}^{m}$ represents the masked FVS of modality $m$, and $ \bm{M}$ masks a portion of the input FVS $\bm{\rho}^{m}$. The auxiliary task intends to reconstruct $\bm{\rho}^{m}$ from $\dot{\bm{\rho}}^{m}$. Since the SA module allows the model to selectively attend to different parts of the input sequence, enhancing its reconstruction capabilities and capturing meaningful dependencies within the data, it serves as a shared component that applies to both the main and auxiliary tasks. The difference between the reconstruction loss for the main and auxiliary tasks lies solely in the input. The main task utilizes the original FVS, whereas the auxiliary task uses the masked FVS to reconstruct the original FVS. The network leverages the knowledge from both tasks and improves the highlight detection performance.

\begin{table*}[b]
\belowrulesep=0pt
\aboverulesep=0pt
\setlength{\tabcolsep}{4.8pt} % Default value: 6pt
\renewcommand{\arraystretch}{1.3} % Default value: 1
\scriptsize
\centering
\caption{Experimental results on the YouTube Highlights (YTH) dataset (mAP)}\label{tab:tab1}
\begin{tabular}{m{1.3cm}<{\centering}|m{1.2cm}<{\centering}m{1.2cm}<{\centering}|m{1.2cm}<{\centering}m{1.2cm}<{\centering}m{1.2cm}<{\centering}m{1.2cm}<{\centering}|m{2.2cm}<{\centering}m{2.7cm}<{\centering}}
\toprule[1pt]

\multirow{4}{*}{\small{Topic}} &\multicolumn{2}{c}{\small{Supervised}}  & \multicolumn{4}{c}{\small{Weakly supervised}} & \multicolumn{2}{c}{\small{Unsupervised}}\\ %\cline{2-9}

~ & \multicolumn{2}{c}{\multirow{1}*{\footnotesize{(frame-level annotations)}}}& \multicolumn{4}{c}{\multirow{1}*{\footnotesize{(video-level annotations)}}} & \multicolumn{1}{c}{\footnotesize{(trained on YTH)}} & \footnotesize{(\textbf{w/o using YTH data})}\\ \cline{2-9}

~ & GIFs & LSVM & RARE & LM & MINI-Net & LR  & CHD & Ours \\ \hline
\footnotesize{dog} & 0.308 & 0.60 & 0.49 & 0.579 & 0.577 & 0.554 & 0.606 & 0.642 \\
\footnotesize{gymnast.} & 0.335 & 0.41 & 0.35 & 0.417 & 0.574 & 0.623 & 0.711 & 0.758 \\ 
\footnotesize{parkour} & 0.540 & 0.61 & 0.50 & 0.670 & 0.698 & 0.701 & 0.742 & 0.709 \\  
\footnotesize{skating} & 0.554 & 0.62 & 0.25 & 0.578 & 0.522 & 0.691 & 0.498 & 0.456 \\  
\footnotesize{skiing} & 0.328 & 0.36 & 0.22 & 0.486 &  0.539 & 0.601 & 0.682 & 0.665 \\  
\footnotesize{surfing} & 0.541 & 0.51 & 0.49 & 0.651 & 0.593 & 0.598 & 0.685 & 0.667 \\  \midrule[1pt] \midrule[1pt]
\footnotesize{Average} & 0.464 & 0.54 & 0.38 & 0.564 &  0.584 & 0.630 & \textbf{0.654} & \underline{0.651} \\  \bottomrule[1pt]
\multicolumn{9}{l}{\footnotesize{The best and the second best overall performances are marked in bold and underlined respectively.}}
\end{tabular}
\end{table*}

\begin{algorithm}[t]
\caption{Pretraining of the proposed method}
\label{alg:A}
\begin{algorithmic}[1]
\STATE {\textbf{Preprocessing:} Initialize the network parameters $\theta$} 
\WHILE{$\theta$ has not converged }
\STATE calculating $\bm{\rho}^{v}$ and $\bm{\rho}^{a}$ from the sampled video visual frames and the audio waves. Randomly generating the mask $\bm{M}$ and obtain $ \dot{\bm{\rho}}^{m}$;
\FOR{$ m = v,a $}
\STATE Compute $\bm{r}^{m}$ based on Eqs. (1), (2), (3);
\STATE Compute $\bm{s}^{m}$ based on Eq. (4);
\STATE Compute $L_{e}^m$ based on Eq. (5);
\STATE Obtain $\Phi^k$ and $\Pi^k$ based on Eqs. (6), (7);
\STATE Compute $L_{r}^m$ based on Eq. (8);
\STATE Compute $L_{au}^m$ based on Eq. (10);
\ENDFOR
\STATE Compute $L_{s}$ based on Eq. (9);
\STATE Compute $L_{total}$ based on Eq. (11);
\STATE Update $\theta$ via gradient back-propagation from $L_{total}$.
\ENDWHILE 
\end{algorithmic}
\end{algorithm}

\subsection{Implementation}
Based on the descriptions of the RASL module, the SCL module and the auxiliary task, the total loss of the proposed framework is shown as follows:
\begin{equation}
\begin{aligned}
\label{eq11}
L_{total} = \sum_{m=v,a}( L_{e}^m + L_{r}^m  + L_{au}^{m}) + L_{s}
\end{aligned}
\end{equation}

\noindent
The main and auxiliary tasks are trained simultaneously. The gradients of the network are backpropagated after the total loss $L_{total}$ is obtained. The pretraining procedure of the proposed method is organized in Algorithm 1. We use the RMSprop optimizer \cite{graves2013generating} with a learning rate of $0.001 $. The mask $\bm{M}$ in Eq. (10) masks $50\%$ feature vectors of the input FVS $\bm{\rho}^{m}$. The feature extractors \cite{liu2021swin} and \cite{liu2020mockingjay} are used to build the visual and the audio FVSs from visual frames and audio waves, respectively. The visual and audio feature extractors contain transformer structures and use attention mechanisms to learn semantics, enabling the generation of high-quality representations. The visual and audio branches of the network have the same architectures. The encoder $E$ and the decoder $D$ contain 3 convolutional and 3 deconvolutional layers, respectively, with a stride of 2. There are ReLU activation functions that follow each convolutional and deconvolutional layer. The hyperparameter $k$ in Eq. (6), (7) and (8) is set to 10. The initial value of $\Gamma $ is set to 3.1. We implement the FVS generation and model pretraining steps sequentially using a server equipped with a 10-core CPU and a Tesla T4 GPU, which has 16 GB of memory capacity. The batch size for model pretraining was 8. We clip the input video length with a fixed length of 30 seconds. When the input or the remaining clipped video length is less than 30 seconds, we concatenate the video repeatedly until it is longer than 30 seconds and clip it. For the visual modality, we sampled the video every 0.2 seconds; thus, the temporal length of the visual FVS was 150. For the audio modality, the temporal density of the feature vectors extracted from audio waves is much greater than that of the visual feature vectors. To reduce the network parameters, we use average pooling to downsample the temporal length to 150. Linear interpolation is used to resample the vector length $ \bm{\rho}_{i}^{a}$ to be equal to $ \bm{\rho}_{i}^{v}$. During inference, if the video is longer than 30 seconds, we first split it into clips of 30 seconds and sequentially input the clips into the networks. Then, we concatenate and normalize the highlight score sequences obtained from the network as the final result. If the input or the clipped video length is shorter than 30 seconds, we repeatedly concatenate and clip the video in the same way as in the training data preparation and then use the portion of the highlight score sequence corresponding to the original video as the result. The network is pretrained with the videos of the large-scale dataset ActivityNet \cite{caba2015activitynet}. The pretrained model can be directly used for inferring highlights of wild videos. Users can determine the highlight score threshold to segment the videos according to the desired highlight length.

\section{Experiments}
In this section, we conduct extensive experiments and comparisons to evaluate the performance of the proposed method. Following other methods \cite{xiong2019less,ye2021temporal}, the experiments were conducted on the popular datasets YouTube Highlights \cite{sun2014ranking} and TVSum \cite{song2015tvsum}. There are various methods for comparison, including supervised, weakly supervised and unsupervised methods.

\subsection{Evaluation datasets}
We evaluate our method on YouTube Highlights and TVSum with the model pretrained on ActivityNet. The YouTube Highlights dataset contains 6 specific categories: surfing, skating, skiing, gymnastics, parkour, and dog. Each category consists of 100 segment-level annotated videos, and the accumulated time is approximately 1,430 minutes.
%Half of the dataset is splitted for testing.
TVSum contains 10 specific categories, including changing vehicle tires, grooming an animal, making sandwiches, parades, flash mob gatherings, and others, with 5 frame-level annotated videos in each category. Following \cite{xiong2019less,panda2017weakly,panda2017collaborative}, we average the importance scores of the manual labels of every segment to achieve segment-level highlight scores.
%and the test part contains the 3 shortest videos of each category. 
To evaluate the proposed method, we report the mean average precision (mAP) and top 5 mAPs for YouTube Highlights and TVSum, respectively, as in \cite{xiong2019less,panda2017weakly,panda2017collaborative}.

\begin{table*}[t] 
\belowrulesep=0pt
\aboverulesep=0pt
\renewcommand{\arraystretch}{1.3} 
\centering
\caption{Experimental results on the TVSum dataset (Top 5 mAP)}\label{tab:tab2}
\scriptsize
\begin{tabular}{c|ccccc|ccccccccc|m{0.9cm}<{\centering} m{1.2cm}<{\centering}}
\toprule[1pt]

\multirow{4}{*}{\small{Topic}} &\multicolumn{5}{c}{\multirow{2}*{\small{Supervised}}}  & \multicolumn{9}{c}{\multirow{2}*{\small{Weakly supervised}}} & \multicolumn{2}{c}{\small{Unsupervised}}\\ %\cline{2-9}
~ & \multicolumn{5}{c}{\multirow{1}*{\small{(frame-level annotations)}}}& \multicolumn{9}{c}{\multirow{1}*{\small{(video-level annotations)}}} & (trained on TVSum) & (\textbf{w/o using TVSum data})\\ \cline{2-17}

~ & KVS & DPP & sLSTM & SM & SMRS & Quasi & MBF &  CVS & SG & VESD & DSN & LM & \makecell[c]{MINI-Net} & LR & CHD & Ours \\ \hline
VT    & 0.353 & 0.399 & 0.411 & 0.415 & 0.272 & 0.336 & 0.295 & 0.328 & 0.423 & 0.447 & 0.373 & 0.559 & 0.785 & 0.850 & \small{-} & 0.897 \\ 
VU &0.441 &0.453 &0.462 &0.467 &0.324 & 0.369 & 0.357 & 0.413 & 0.472 & 0.493 & 0.441 & 0.429 & 0.566 & 0.714 & \small{-} & 0.657 \\ 
GA & 0.402 &0.457 &0.463 &0.469 &0.331 & 0.342 & 0.325 & 0.379 & 0.475 & 0.496 & 0.428 & 0.612 & 0.736 & 0.819 & \small{-} & 0.925 \\ 
MS & 0.417 & 0.462 & 0.477 & 0.478 & 0.362 & 0.375 & 0.412 & 0.398 & 0.489 & 0.503 & 0.436 & 0.540 & 0.753 & 0.786 & \small{-} & 0.583 \\ 
PK        & 0.382 & 0.437 & 0.448 & 0.445 & 0.289 & 0.324 & 0.318 & 0.354 & 0.456 & 0.478 & 0.411 & 0.604 & 0.769 & 0.802 & \small{-} & 0.847 \\ 
PR        & 0.403 & 0.446 & 0.461 & 0.458 & 0.276 & 0.301 & 0.334 & 0.381 & 0.473 & 0.485 & 0.417 & 0.475 & 0.633 & 0.755 & \small{-} & 0.830 \\ 
FM     & 0.397 & 0.442 & 0.452 & 0.451 & 0.302 & 0.318 & 0.365 & 0.365 & 0.464 & 0.487 & 0.412 & 0.432 & 0.612 & 0.716 & \small{-} & 0.697 \\ 
Bk      & 0.342 & 0.395 & 0.406 & 0.407 & 0.297 & 0.295 & 0.313 & 0.326 & 0.417 & 0.441 & 0.368 & 0.663 & 0.756 & 0.773 & \small{-} & 0.833 \\ 
BT     & 0.419 & 0.464 & 0.471 & 0.473 & 0.314 & 0.327 & 0.365 & 0.402 & 0.483 & 0.492 & 0.435 & 0.691 & 0.756 & 0.786 & \small{-} & 0.875 \\ 
DS        & 0.394 & 0.449 & 0.455 & 0.453 & 0.295 & 0.309 & 0.357 & 0.378 & 0.466 & 0.488 & 0.416 & 0.626 & 0.656 & 0.681 & \small{-} & 0.890 \\ \midrule[1pt] \midrule[1pt]
Average         & 0.398 & 0.447 & 0.451 & 0.461 & 0.306 & 0.329 & 0.345 & 0.372 & 0.462 & 0.481 & 0.424 & 0.563 & 0.702 & \underline{0.768} & 0.528 & \textbf{0.783} \\  \bottomrule[1pt]
\multicolumn{17}{l}{\footnotesize{The best and the second best overall performances are marked in bold and underlined, respectively.}}
\end{tabular}
\end{table*}

\subsection{Comparisons}
We compare our proposed framework to other competing approaches for evaluation. The compared supervised methods include LSVM \cite{sun2014ranking}, GIFs \cite{gygli2016video2gif}, KVS \cite{potapov2014category}, sLstm \cite{zhang2016video}, DPP \cite{gong2014diverse}, SMRS \cite{elhamifar2012see} and SM \cite{gygli2015video}. The compared weakly supervised approaches include RARE \cite{yang2015unsupervised}, MBF \cite{chu2015video}, CVS \cite{panda2017collaborative}, DSN \cite{panda2017weakly}, VESD \cite{cai2018weakly}, SG \cite{mahasseni2017unsupervised}, Quasi \cite{kim2014joint}, LM \cite{xiong2019less}, MINI-Net \cite{hong2020mini}, and LR \cite{ye2021temporal}, and the compared unsupervised method is CHD \cite{badamdorj2022contrastive}. Among them, LR and MINI-Net also utilize visual and audio modalities. CHD is another unsupervised method that does not require video labels. In addition to highlight detection methods, several of these methods involve video summarization approaches. Since most of the important frames detected belong to highlights, we still compare these summarization approaches using the same metrics. The results of the compared methods evaluated on the two datasets are reported in \cite{ye2021temporal,xiong2019less,badamdorj2022contrastive}.

\begin{table}[b]
\setlength{\tabcolsep}{4.8pt} % Default value: 6pt
\renewcommand{\arraystretch}{1.3} % Default value: 1
\footnotesize
\centering
\caption{The ablation study results on the datasets}\label{tab:tab3}
\begin{tabular}{p{2.3cm}<{\centering}c p{2.3cm}<{\centering}}
\toprule[1pt]
Method & YouTube Highlights & TVSum \\ \midrule[1pt]
Ours w/o RASL &  0.560 & 0.616 \\
Ours w/o SA &  0.638 & 0.711 \\
Ours w/o auxiliary & 0.644 & 0.724 \\ 
Ours w/o vision & 0.625 & 0.614 \\
Ours w/o audio & 0.630 & 0.704 \\ 
Ours & \textbf{0.651} & \textbf{0.783} \\ \bottomrule[1pt]

\end{tabular}
\end{table}

\begin{table}[b]
\setlength{\tabcolsep}{4.8pt} % Default value: 6pt
\renewcommand{\arraystretch}{1.3} % Default value: 1
\footnotesize
\centering
\caption{The comparisons of different audio-visual fusion schemes}\label{tab:tab4}
\begin{tabular}{ccp{2.5cm}<{\centering}}
\toprule[1pt]
Method & YouTube Highlights & TVSum \\ \midrule[1pt]
Summation &  0.630 & 0.700  \\ 
Concatenation &  0.628 & 0.716 \\ 
Submodule MLP \cite{hong2020mini} &  0.589 & 0.604 \\ 
low-rank fusion \cite{ye2021temporal} &  0.621 & 0.736\\ 
Ours &  \textbf{0.651} & \textbf{0.783} \\ \bottomrule[1pt]

\end{tabular}
\end{table}

\begin{figure}[t]
  \centering
   \includegraphics[width= 9 cm]{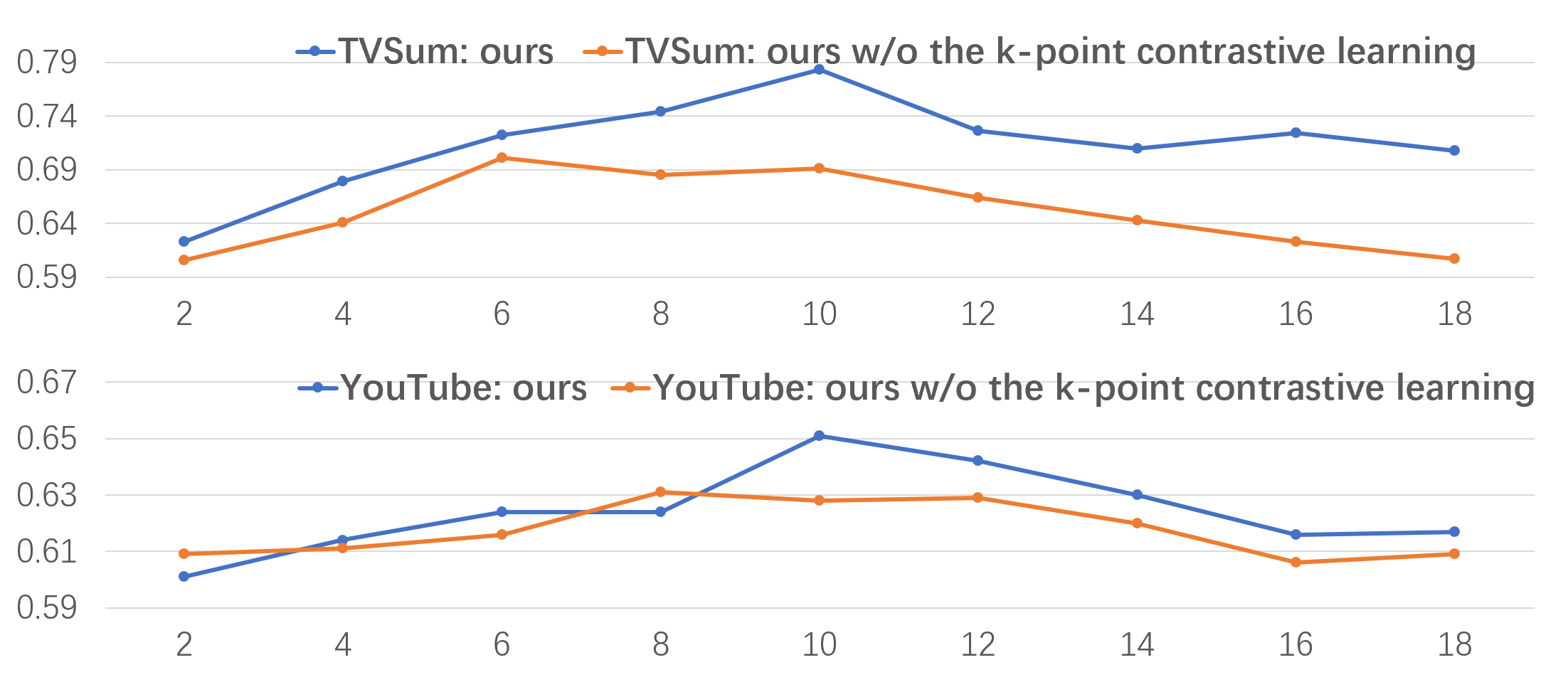}
  \caption{The performance (mAP) varies by changing the {\itshape k} value of the {\itshape k}-point contrastive learning.
 }
\end{figure}

\begin{figure*}[tp]
  \centering
  \includegraphics[width=15 cm]{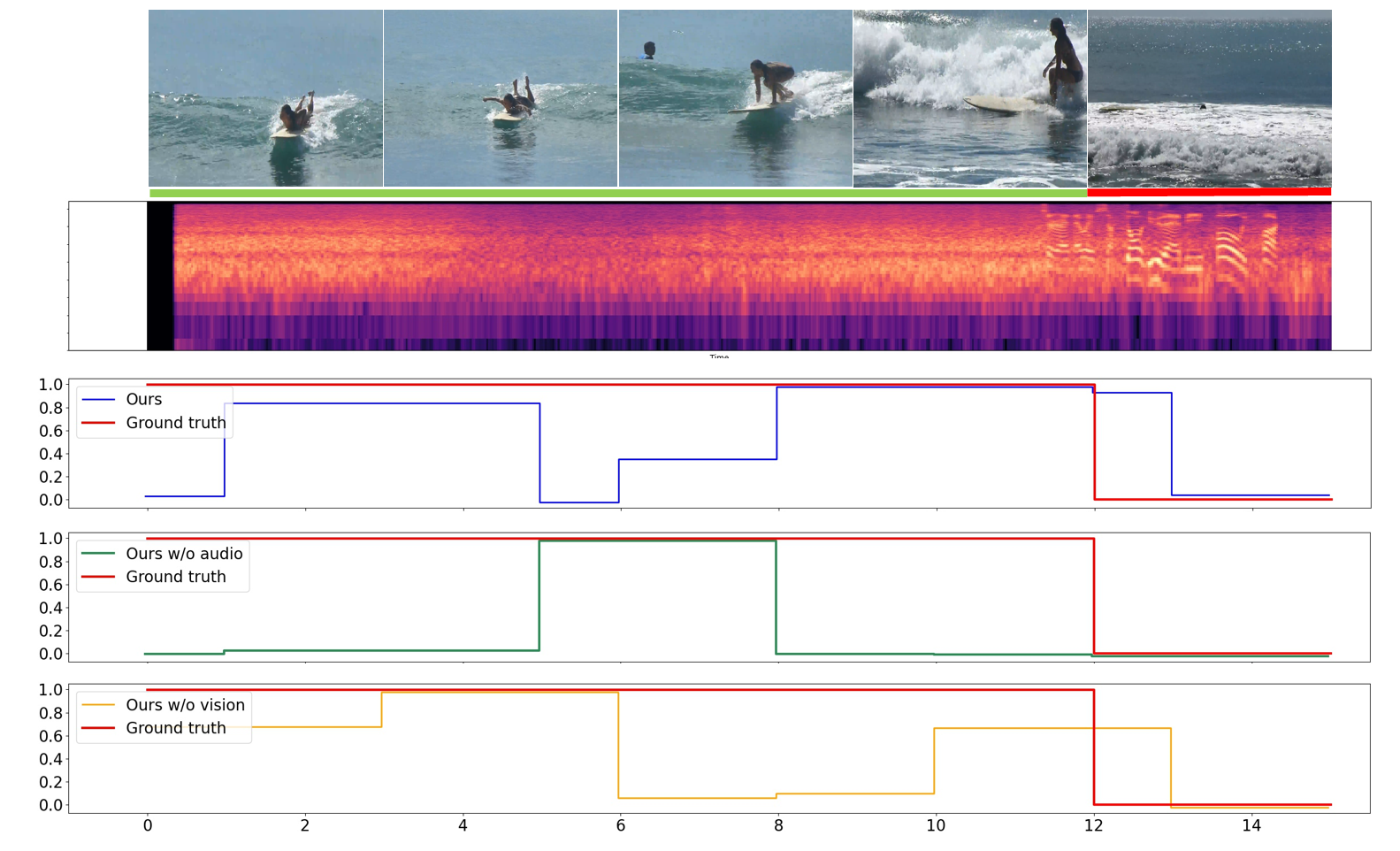}
  \caption{Illustration of the SCL module that improves the highlight detection performance. The subfigures in the first and the second rows are the representative visual frames and the audio mel-spectrogram of the input video. The subfigures in the third, fourth and last rows are the highlight score curves estimated by the fully proposed method, the proposed method pretrained without the audio modality and the method trained without the visual modality, respectively. The proposed method pretrained without the audio modality leaves out the beginning and the tail parts, while the proposed method pretrained without the visual modality leaves out the middle part of the highlights. In contrast, the full proposed method with the SCL module connecting the two modalities yields the best overall results.
 }
\end{figure*}

\begin{figure*}[h]
  \centering
  \includegraphics[width= 12 cm]{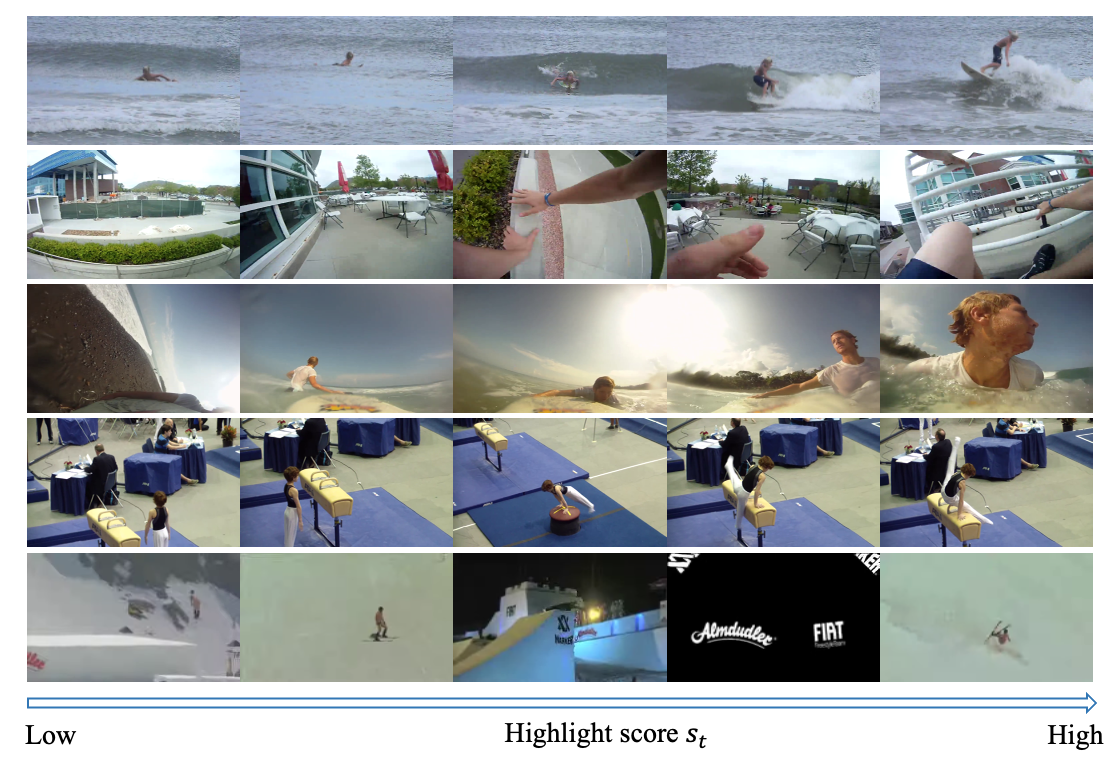}
  \caption{The example highlighting the prediction ability of the proposed method on YouTube Highlights. The highlight score $s_t $ increases along the axis from the left to the right.
 }
\end{figure*}

Table I shows the comparison results on YouTube Highlights. Compared to the weakly supervised approaches MINI-Net and LR, which also use visual and audio modalities, our unsupervised method achieves average gains of 6.7\% and 2.1\%, respectively. Notably, our method infers the highlight segments without requiring the audio modality as the input, while MINI-Net and LR require both visual frames and audio waves as the input. Our method also achieves better average performance than do the supervised methods LSVM and GIFs. This is because supervised methods easily fall into the problem of overfitting frame-level data. Our method and CHD demonstrated very similar overall performances. However, CHD is trained on the YouTube Highlights training split. It leverages the inclusion of similar videos in both the training and test splits, allowing it to easily align with the test video distribution. In contrast, our method trains on an unrelated dataset. This allows our method to be directly applicable and practical for real-world applications, especially in scenarios where data collection is laborious or data availability is constrained. We also find that the proposed method shows relatively weak performance in terms of the skating video category. This is because the cameras of most skating videos of YouTube Highlights shake severely and do not face skaters stably, which is different from the usual visual presentation forms in the unlabeled pretraining dataset ActivityNet. This deteriorates the performance of our method for skating videos.
Table II shows the experimental results on TVSum. TVSum consists of 10 categories of diverse long videos. As another unsupervised method, CHD, only reports its overall performance on TVSum but not across TVSum categories \cite{badamdorj2022contrastive}; we only compare its overall performance. The proposed method achieves the best overall performance compared to all the competing methods. We also observe that the proposed method outperforms CHD with a large gain of 25.5\%, even though CHD is trained on the TVSum training split. This shows the robust and adaptive performance of the proposed method. We find that the proposed method shows relatively weak performance in the making sandwich (MS) category. This is because videos in this category include segments where chefs use exaggerated statements and actions to introduce the sandwich and capture the attention of viewers. However, these segments are not considered key segments based on prior knowledge of the topic of making sandwiches. In contrast, the weakly supervised trained methods MINI-Net and LR learn highlight segments given the video topics, resulting in better performance for videos in the sandwich category. Overall, the unsupervised proposed method achieves 8.1\% and 1.5\% gains in terms of the average score by the most competitive methods, MINI-Net and LR, respectively. This demonstrates the superior performance of the proposed method.

\subsection{Ablation study}
In this subsection, we conduct ablation studies to investigate the effects of the proposed components in the model. We also evaluate the impacts of the modality fusion schemes and the values of {\itshape k} in the {\itshape k}-point contrastive learning and the reconstruction targets.

\noindent {\bf{Impacts of the model components:}}  We evaluate the model components and show the ablation study results in Table III. We first evaluate the effect of the proposed RASL module, which uses {\itshape k}-point contrastive learning. In this case, we set the value of {\itshape k} to 0, and the loss term $ L_{r}^m$ is excluded. Without the {\itshape k}-point contrastive learning, the RASL module degrades to a reweighting module. The performance of the proposed method drops by 9.1\% and 16.7\% on YouTube Highlights and TVSum, respectively. This finding certifies that the proposed RASL module using {\itshape k}-point contrastive learning can emphasize important representation activations and improve highlight detection performance. We also drop the SA module to evaluate the effect of SA enhancement on FVS. Without the SA module, the performance drops by 2.3\% and 7.2\% on YouTube Highlights and TVSum, respectively. This finding certifies that the SA module can improve the representation performance of FVS. We also ablate the auxiliary masked FVS reconstruction task to evaluate the effect of the MTL on our model learning. In this case, the proposed method without MTL also falls behind the full version. The results indicate that the MTL framework improves highlight detection performance by learning knowledge across the main and auxiliary tasks. We also show the experimental results of pretraining the model using only a single modality. The performance degradation indicates that the association between the paired visual and audio perceptions in the pretrained model improves the performance.

\begin{table}[t]
\belowrulesep=0pt
\aboverulesep=0pt
\setlength{\tabcolsep}{4.8pt} % Default value: 6pt
\renewcommand{\arraystretch}{1.3} % Default value: 1
\footnotesize
\centering
\caption{The comparisons of reconstruction targets}\label{tab:tab5}
\begin{tabular}{cccp{1.2cm}<{\centering}}
\toprule[1pt]
Visual target& Audio target & YouTube Highlights & TVSum  \\ \midrule[1pt]
Pixel &   mel-spectrogram & 0.639 & 0.603 \\ 
Pixel &  Audio FVS \cite{liu2020mockingjay} & 0.648 & 0.608\\ 
Visual FVS \cite{liu2021swin} &  mel-spectrogram & 0.649 & 0.700 \\ 
Visual FVS \cite{liu2021swin} &  Audio FVS\cite{liu2020mockingjay} & \textbf{0.651} & \textbf{0.783}\\ 
\bottomrule[1pt]
\end{tabular}
\end{table}

\noindent {\bf{Impact of the modality fusion scheme:}} We compare 4 other modality fusion schemes to our proposed SCL module. Summation and concatenation are the two most commonly used approaches for modal fusion. Submodule MLP and low-rank fusion are two modal fusion schemes proposed in the highlight detection methods \cite{hong2020mini} and \cite{ye2021temporal}, respectively. As all the compared fusion schemes directly fuse the required visual and audio modalities, we fuse the self-attended FVS $\bar{\bm{\rho}}^{v}$ and $ \bar{\bm{\rho}}^{a}$ using the compared fusion schemes and send the fused FVS to a network with the same architecture as one branch of the proposed network for self-reconstruction. Table IV shows the comparison results. Our method achieves the best performance with the proposed SCL module. This is because some sounds of the attractive scenes in YouTube Highlights and TVSum are disrupted by irrelevant environmental sounds or background music, which decreases the performance of the above direct fusion schemes. The submodule MLP even suffers from overfitting, as the complex network has a large number of parameters. Compared to other schemes that directly fuse the representations of the two modalities, the proposed model trained employing the SCL module requires only the visual modality as the input during inference. The visual branch of our model generates the highlight score sequence by inferring representation vectors with paired visual-audio semantics without the probable audio noise, as the overall correct connection of the related visual and audio semantics has been established by the SCL module during pretraining on ActivityNet, where the existing audio noise varies and does not consistently lead the model to converge toward a fixed erroneous modal connection.

\noindent {\bf{Impact of the {\itshape k}-point contrast learning:}} In addition, we investigate the effect of the {\itshape k}-point contrast learning and the value {\itshape k} on our framework. We ablate the contrast learning by using $ \eta = \frac{1}{k} \sum_{i=0}^{k} \hat{s}^{m}_{i}$ as a substitute in Eq. (8) for comparison. Fig. 6 shows the results of our framework with and without {\itshape k}-point contrastive learning with varying values of {\itshape k}. The performance of the proposed method with {\itshape k}-point contrastive learning is better overall than that without {\itshape k}-point contrast learning. Notably, the gaps between the curves are relatively larger when the value of {\itshape k} is high than when {\itshape k} is small. This is because our {\itshape k}-point contrastive learning suppresses noisy activations when using a large {\itshape k}.

\begin{table}[t]
\belowrulesep=0pt
\aboverulesep=0pt
\setlength{\tabcolsep}{4.8pt} % Default value: 6pt
\renewcommand{\arraystretch}{1.3} % Default value: 1
\footnotesize
\centering
\caption{The comparisons of the efficiency}\label{tab:tab6}
% \begin{tabular}{|c c|p{1.5cm}<{\centering}|p{1.5cm}<{\centering}|}
\begin{tabular}{m{1.7cm}<{\centering}m{2.3cm}<{\centering}m{1.7cm}<{\centering}m{1.7cm}<{\centering}}
\toprule[1pt]
Method & Training method & Parameters & FLOPs      \\ \midrule[1pt]
GIFS & Supervised & 440.31M & 823.44G  \\ 
sLSTM & Supervised & 2.69M & 0.07G   \\ 
SG &  Weakly supervised & 18.78M & 9.16G  \\ 
Ours  & Unsupervised & 8.53 M & 0.82G \\ \bottomrule[1pt]
\end{tabular}
\end{table}

\noindent {\bf{Impact of the reconstruction targets}}: We also demonstrated that the use of FVS as the reconstruction target is more beneficial for highlight detection than the use of visual pixels and audio mel-spectrograms. Table V shows comparisons of the proposed model trained on different reconstruction targets for highlight detection. The results in Table V demonstrate that utilizing the visual and audio FVSs extracted by deep networks \cite{liu2021swin} and \cite{liu2020mockingjay} as the reconstruction targets achieves the best performance for highlight detection. Replacing one or both of the deep feature extractors with visual pixels or mel-spectrograms results in a degradation in performance. This is because features extracted by deep neural networks have better representations than traditional pixels and mel-spectrograms. It is also possible that the performance of the proposed highlight detection method can be further improved with other advanced features due to the rapid development of deep feature extractors. Notably, our results using very simple feature pixels and a mel-spectrogram still outperform 6 of the 7 methods in Table I and 13 of the 15 methods in Table II for YouTube Highlights and TVSum, respectively, even other methods use complex features. This demonstrates the robustness of the proposed framework to different input features.

\subsection{Qualitative evaluation}
This subsection presents the qualitative evaluation results for further study. In Fig. 7, we show representative frames, an audio mel-spectrogram of an input video and the estimated highlight score curves. The highlight score curves are estimated using three models: the proposed network, the proposed network pretrained without using the audio modality, and the proposed network pretrained without using the visual modality. The proposed method, pretrained without the audio modality, yields high estimated scores in the middle, where the surfer strongly moves from lying on the surfboard to standing on the surfboard. However, this approach ignores the beginning and the tail highlight parts, where the surfer is motionless, either lying or standing on the surfboard. The proposed method, pretrained without the visual modality, covers a longer range of correct high highlight scores. However, relatively low scores were observed in the middle, where degradation of the mel-spectrogram was observed. It is worth noting that both our method pretrained without the audio and the visual modalities give no credit at the end of the video. This is because, visually, there are only sea waves and no surfing motion at the tail. Acoustically, the pattern of the mel-spectrogram at the tail was different from that of the overall mel-spectrogram. With the SCL module, the proposed method considering cross-modal semantics yields the best overall results.

We also demonstrate the representative frames from five highlight detection results on YouTube Highlights in Fig. 8. In the first four instances, we find that the representative frames with higher values of $s_t $ represent attractive segments that contain more information, while the frames with low $s_t$ values are less meaningful. This indicates that our unsupervised method effectively identifies highlight moments. However, in the last instance, where abrupt video transitions occur, the network mistakenly identifies these transitions as highlight frames. Although these sudden transitions draw viewer attention, they are not relevant to the video topic and hence should not be considered highlights. This shows the limitation of our proposed method, as it might classify some irrelevant frames that capture attention as highlights.

\subsection{Efficiency evaluation}
We also conduct an efficiency evaluation of the proposed method and compare it to two supervised methods and one weakly supervised method that have available codes. As different official implementation platforms with various efficiencies are used, we count the number of parameters and floating point operations (FLOPs) for independently assessing the computational complexity of each algorithm. Considering that the inputs for all the methods are sustainable video features, we ensure a fair comparison by excluding the feature extractors and focusing solely on the network architectures for all the methods. The experimental results are shown in Table VI. The supervised method GIFs has the highest number of parameters and FLOPs, while the sLSTM has the lowest. Notably, the proposed method also yields a low FLOP value, specifically lower than 1G, and has half the number of parameters compared to SG.

\section{Conclusion}
In this paper, we present a novel unsupervised cross-modal highlight detection framework. We propose the RASL module with the {\itshape k}-point contrastive learning mechanism to learn the significant activations through a self-reconstruction task. To enable the network to connect the visual and audio modalities, we propose the SCL module to learn paired representations. Given only the visual frames, the cross-modal pretrained network can generate representations with visual-audio-level semantics and directly infer the highlight scores. An auxiliary task of masked FVS reconstruction is used to enhance the representation. The experimental results demonstrate the effectiveness and superior performance of the proposed approach compared to other state-of-the-art highlight detection approaches.

\bibliographystyle{IEEEtran}
% \bibliography{sample-base}

\newpage
\vfill

\end{document}